\documentclass[sigconf]{acmart}

\usepackage[inline]{enumitem} 
\setlist[itemize]{leftmargin=*, nosep}
\usepackage{multirow}
\usepackage{bbding} 
\usepackage{acmart-taps}
\usepackage{acronym}
\acrodef{AI}{artificial intelligence}
\acrodef{LLM}{large language model}
\acrodef{VLM}{vision-language model}
\acrodef{AR}{augmented reality}
\acrodef{TACC}{theme accuracy}
\acrodef{EACC}{entity accuracy}
\acrodef{IOU}{intersection over union}
\acrodef{TNR}{true negative rate}
\acrodef{FPR}{false positive rate}
\acrodef{PEFT}{parameter-efficient fine-tuning}
\acrodef{LoRA}{low rank adaptation}

\newcommand{\OurData}[1]{\textsc{LEGO-VLM}}

\newcommand{\jp}[1]{#1}

\newcommand{\subhead}[1]{\noindent{\textbf{#1}}}

\copyrightyear{2026}
\acmYear{2026}
\setcopyright{cc}
\setcctype{by-nc-nd}
\acmConference[ICMI '26]{INTERNATIONAL CONFERENCE ON MULTIMODAL INTERACTION}{October 05--09, 2026}{Napoli, Italy}
\acmBooktitle{INTERNATIONAL CONFERENCE ON MULTIMODAL INTERACTION (ICMI '26), October 05--09, 2026, Napoli, Italy}
\acmDOI{10.1145/3776574.3831119}
\acmISBN{979-8-4007-2318-6/2026/10}

\acmSubmissionID{1093}

\begin{document}

\title{LEGO Co-builder: Exploring Fine-Grained Vision-Language Modeling for Multimodal Assembly Assistants}


\author{Haochen Huang}
\authornote{First co-authors.}
\affiliation{%
  \institution{Vrije University Amsterdam}
  \country{Amsterdam, Netherlands}
}
\affiliation{%
  \institution{University of Amsterdam}
  \country{Amsterdam, Netherlands}
}
\affiliation{
  \institution{Centrum Wiskunde \& Informatica}
  \country{Amsterdam, Netherlands}
}
\email{peterdavinci@yahoo.com}

\author{Yue Su}
\authornotemark[1]
\affiliation{%
  \institution{Vrije University Amsterdam}
  \country{Amsterdam, Netherlands}
}
\email{y.su@vu.nl}

\author{Xin Sun}
\affiliation{%
  \institution{National Institute of Informatics}
  \country{Tokyo, Japan}
}
\affiliation{%
  \institution{University of Amsterdam}
  \country{Amsterdam, Netherlands}
}
\email{xsun@nii.ac.jp}

\author{Moonisa Ahsan}
\affiliation{%
  \institution{Centrum Wiskunde \& Informatica}
  \country{Amsterdam, Netherlands}
}
\email{moonisa@cwi.nl}

\author{Mohammad Aliannejadi}
\affiliation{%
  \institution{University of Amsterdam}
  \country{Amsterdam, Netherlands}
}
\email{m.aliannejadi@uva.nl}

\author{Irene Viola}
\affiliation{%
  \institution{Centrum Wiskunde \& Informatica}
  \country{Amsterdam, Netherlands}
}
\email{Irene.Viola@cwi.nl}

\author{Zhaochun Ren}
\affiliation{%
  \institution{Leiden University}
  \country{Leiden, Netherlands}
}
\email{z.ren@liacs.leidenuniv.nl}

\author{Chuang Yu}
\affiliation{%
  \institution{University College London}
  \country{London, United Kingdom}
}
\email{chuang.yu@ucl.ac.uk}

\author{Aneta Lisowska}
\affiliation{%
  \institution{Vrije University Amsterdam}
  \country{Amsterdam, Netherlands}
}
\email{a.j.lisowska@vu.nl}

\author{Artem Belopolsky}
\affiliation{%
  \institution{Vrije University Amsterdam}
  \country{Amsterdam, Netherlands}
}
\email{a.belopolskiy@vu.nl}

\author{Koen Hindriks}
\affiliation{%
  \institution{Vrije University Amsterdam}
  \country{Amsterdam, Netherlands}
}
\email{k.v.hindriks@vu.nl}

\author{Pablo Cesar}
\affiliation{%
  \institution{Centrum Wiskunde \& Informatica}
  \country{Amsterdam, Netherlands}
}
\affiliation{%
  \institution{Technische Universiteit Delft}
  \country{Delft, Netherlands}
}
\email{p.s.cesar@cwi.nl}

\author{Junxiao Wang}
\affiliation{%
  \institution{Guangzhou University}
  \country{Guangzhou, China}
}
\email{junxiao.wang@gzhu.edu.cn}

\author{Jiahuan Pei}
\authornote{Corresponding author.}
\affiliation{%
  \institution{Vrije University Amsterdam}
  \country{Amsterdam, Netherlands}
}
\email{j.pei2@vu.nl}

\renewcommand{\shortauthors}{Huang, Su, and Pei, et al.}

\begin{abstract}
Vision-language models (VLMs) are facing the challenges of understanding and following multimodal assembly instructions, particularly when fine-grained spatial reasoning and precise object state detection are required. In this work, we explore \textbf{LEGO Co-builder}, a hybrid benchmark combining real-world LEGO assembly logic with programmatically generated multimodal scenes. The dataset captures stepwise visual states and procedural instructions, allowing controlled evaluation of instruction-following, object detection, and state detection. 
We introduce a unified framework and assess leading VLMs such as GPT-4o, Gemini, and Qwen-VL, under zero-shot and fine-tuned settings. We also evaluated the framework using a reasoning-focused model, GLM-4.1-thinking. 
Our results show that while object detection achieved high performance (98.16\% with fine-tuned InstructBLIP), fine-grained scene understanding and assembly state detection remain challenging: Fine-tuned MiniGPT-v2 reached only 37.52\% F1 for identifying theme entities, and even advanced models such as GPT-4o achieved just 40.54\% F1 on state detection.
This highlights gaps in fine-grained visual understanding among existing models.
We release the benchmark, codebase, and generation pipeline to support future research on multimodal assembly assistants grounded in real-world workflows.\footnote{\url{https://github.com/peterhuang-coding/LEGO-Co-builder}}
\end{abstract}


\maketitle

\section{Introduction}

Multimodal instruction-following assistants are gaining increasing relevance in domains requiring precise procedural understanding, such as furniture construction~\citep{you2022human}, automotive manufacturing~\citep{bellalouna2020fiaar}, and industrial product assembly~\citep{funk2017working}. These tasks demand step-by-step reasoning, spatial awareness, and accurate interpretation of visual and textual instructions—capabilities that current AI systems still struggle to reliably deliver.

\begin{figure}[t!]
    \centering
    \includegraphics[width=0.95\columnwidth]{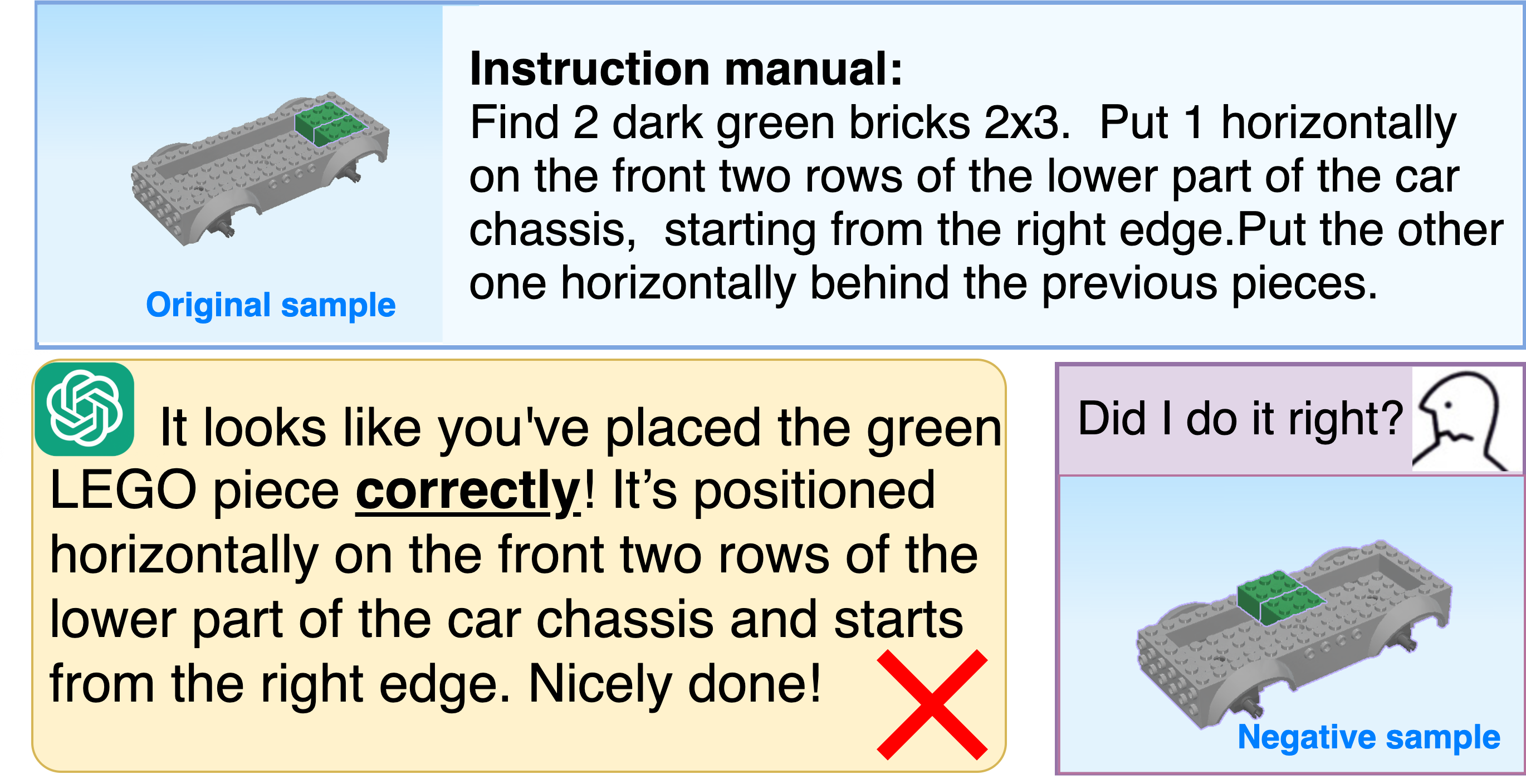}
    \caption{An example of incorrect state detection by GPT-4o during LEGO assembly.
    Given an image and text instructions, the model fails to accurately recognize an incorrectly placed LEGO part (highlighted in green), demonstrating the challenge of fine-grained vision-language alignment.}
    \label{fig:Motivation}
\end{figure}

\begin{figure*}[t!]
\centering
\includegraphics[width=2\columnwidth]{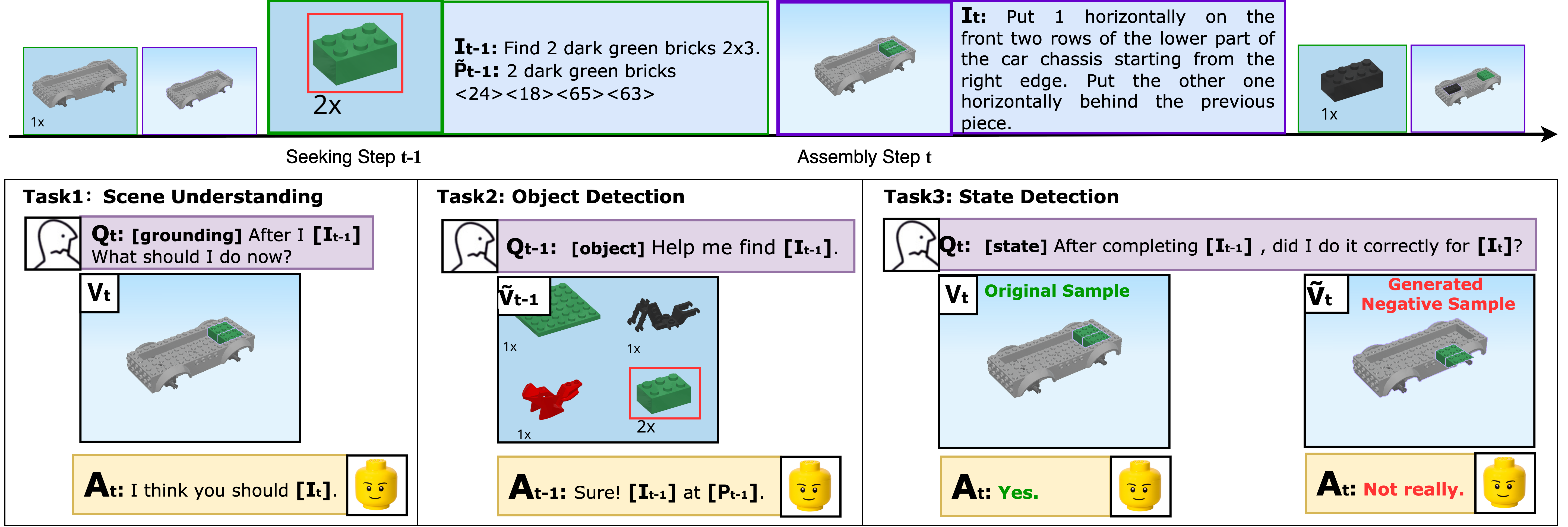}
\caption{An illustration of the proposed three \ac{VLM} tasks, highlighting the core capabilities required by the one-for-all \ac{VLM} architecture for LEGO assembly based on the procedural instruction manual.}
\label{fig:task}
\end{figure*}

Traditional computer vision research has tackled various purely visual tasks, including action segmentation, recognition, anticipation, and object state detection~\citep{Damen2018EPICKITCHENS, miech20rareact, wang2023holoassist, gao2024physically}. Procedural activity understanding has also been explored using video datasets with or without textual narration~\citep{tang2019coin, zhukov2019crosstask, li2023mimic, sener2022assembly101}. Some methods incorporate textual alignment, such as narration-video grounding or instruction-conditioned generation~\citep{padmakumar2021teach, miech2019howto100m}, but often rely on separate architectures for each task type.

\Acfp{VLM} offer a promising solution by bridging visual and textual modalities, enabling models to align procedural language with visual observations. Recent advances in large-scale models such as GPT-4o~\citep{openai2024gpt4v}, LLaVa~\citep{liu2023llava}, Qwen-VL~\citep{bai2023qwen}, BLIP-2~\citep{li2023blip2}, MiniGPT-v2~\citep{chen2023minigpt}, and more recent reasoning-augmented variants like VLM-R1~\citep{shen2025vlm}, GLM-4.1v-thinking~\citep{hong2025glm}, and Gemini2.5~\citep{comanici2025gemini}, have demonstrated strong performance on a range of general-purpose multimodal tasks. 
While most existing benchmarks focus on coarse-grained capabilities like object recognition and captioning, they often fail to evaluate the detailed, procedural understanding required for step-by-step manual assembly.
While recent \acp{VLM} research~\citep{laurenccon2024matters,cheng2024spatialrgpt,chen2024spatialvlm} aim to unify vision-language reasoning under a single architecture, challenges remain—particularly for tasks that require dense and fine-grained perception~\citep{wei2024vary,rahmanzadehgervi2024vision, chen2024we}. 

In this work, we address the need for instruction-following benchmarks that evaluate a model’s ability to reason over both visual and linguistic input in detail. We focus on LEGO brick assembly as a representative manual task that is richly structured, visually complex, and procedurally grounded. 
\autoref{fig:Motivation} illustrates the limitations of current models in such tasks. When prompted with a visual scene and corresponding instruction, GPT-4o fails to detect a misplaced LEGO piece, highlighting the challenge of fine-grained visual scene understanding and object state recognition. This motivates the need for benchmarks that go beyond high-level visual understanding and evaluate precise spatial compliance with procedural steps.

To address this gap, we present \textbf{LEGO Co-builder}, a benchmark and dataset designed to evaluate fine-grained instruction-following capabilities in LEGO assembly tasks. Our dataset includes multimodal sequences consisting of visual snapshots, object states, and procedural steps derived from human-crafted manuals. We introduce a unified task formulation and evaluate nine leading VLMs in both zero-shot and fine-tuned settings. 
Our contributions are summarized as follows:
\begin{enumerate*}[label=(\arabic*)]
\item We develop a unified vision-language architecture to benchmark fine-grained, multimodal, instruction-following capabilities in procedural manual-guided assembly tasks.
\item We evaluate prevailing VLMs on a dataset of LEGO assembly sequences with grounded visual and textual supervision, under both zero-shot and fine-tuned settings.
\item We release the dataset, benchmarking code, and a modular synthetic data generation pipeline to support future research in multimodal instruction-following.
\end{enumerate*}

\textbf{Societal Impact.}
This work advances the development of multimodal AI assistants that can transform how people learn and perform complex physical tasks. 
By enabling fine-grained vision-language understanding, our benchmark supports the creation of multimodal educational tools that combine visual input, language, and potentially \ac{AR} to enhance hands-on learning experiences. 
These systems can improve training in domains like education, surgical practice, and industrial training. 
Moreover, the ability to interpret and verify procedural steps visually opens up promising avenues for assistive technologies—particularly for blind or visually impaired learners—by providing real-time, AI-driven guidance through tasks that were previously inaccessible. 
In this way, our research supports both innovation in teaching modalities and greater inclusivity in skill development.

\section{Fine-Grained Vision-Language Modeling}\label{sec:task-definiton}
\subsection{Task Definition}

We investigate manual-guided LEGO assembly and define the following three vision-language tasks (See in \autoref{fig:task}):

\aptLtoX[graphic=no,type=html]{\begin{enumerate}
    \item[\textbf{(T1)}] \textbf{Scene Understanding}. 
    Given the current step's image $V_{t}$ and the task-specific query $Q_{t}$, the model outputs a response containing the scene description for the current assembly step $I_{t}$.
    $Q_{t}$ is the task description contextualized with the previous step's textual instruction $I_{t-1}$.
    It assesses the model's ability to accurately describe the current assembly step from an instruction manual, including but not limited to generating object representations, their properties, and the assembly procedure.
   \item[\textbf{(T2)}] \textbf{Object Detection}. 
    Given a seeking step's image $V_{t'}$ ($t'=t-1$) and task-specific query $Q_{t'}$, the model outputs a response containing
    the positional text $P_{t'}$, which includes identified object and its coordinate formatted as ``[Object] [Xleft] [Ytop] [Xright] [Ybottom]''. 
    The query $Q_{t'}$ is the task description contextualized with the corresponding textual instruction $I_{t'}$.
    It assesses the model's ability to accurately identify and locate objects in the scene, ensuring they are positioned correctly for the next assembly action. 
   \item[\textbf{(T3)}] \textbf{State Detection}. 
    Given an assembly step's image, either the original manual image $V_{t}$ or a generated negative sample $\tilde{V}_{t}$, and task-specific query $Q_{t}$, the model outputs a response indicating a correct or incorrect state.
    It evaluates the model's ability to verify the accuracy of the assembly progression, determining if the assembly action has been correctly completed.
\end{enumerate}}{\begin{enumerate}[label=(\textbf{T\arabic*}), nosep, leftmargin=*, itemindent=2.2em]
    \item \textbf{Scene Understanding}. 
    Given the current step's image $V_{t}$ and the task-specific query $Q_{t}$, the model outputs a response containing the scene description for the current assembly step $I_{t}$.
    $Q_{t}$ is the task description contextualized with the previous step's textual instruction $I_{t-1}$.
    It assesses the model's ability to accurately describe the current assembly step from an instruction manual, including but not limited to generating object representations, their properties, and the assembly procedure.
   \item \textbf{Object Detection}. 
    Given a seeking step's image $V_{t'}$ ($t'=t-1$) and task-specific query $Q_{t'}$, the model outputs a response containing
    the positional text $P_{t'}$, which includes identified object and its coordinate formatted as ``[Object] [Xleft] [Ytop] [Xright] [Ybottom]''. 
    The query $Q_{t'}$ is the task description contextualized with the corresponding textual instruction $I_{t'}$.
    It assesses the model's ability to accurately identify and locate objects in the scene, ensuring they are positioned correctly for the next assembly action. 
    \item \textbf{State Detection}. 
    Given an assembly step's image, either the original manual image $V_{t}$ or a generated negative sample $\tilde{V}_{t}$, and task-specific query $Q_{t}$, the model outputs a response indicating a correct or incorrect state.
    It evaluates the model's ability to verify the accuracy of the assembly progression, determining if the assembly action has been correctly completed.
\end{enumerate}}
These tasks have been crafted to rigorously test and demonstrate the capabilities of \acp{VLM} in LEGO Assembly, focusing on their ability to understand and interpret complex multimodal inputs.

\subsection{Unified Vision-Language Architecture}
\begin{figure}[t!] 
\centering 
\includegraphics[width=1\columnwidth]{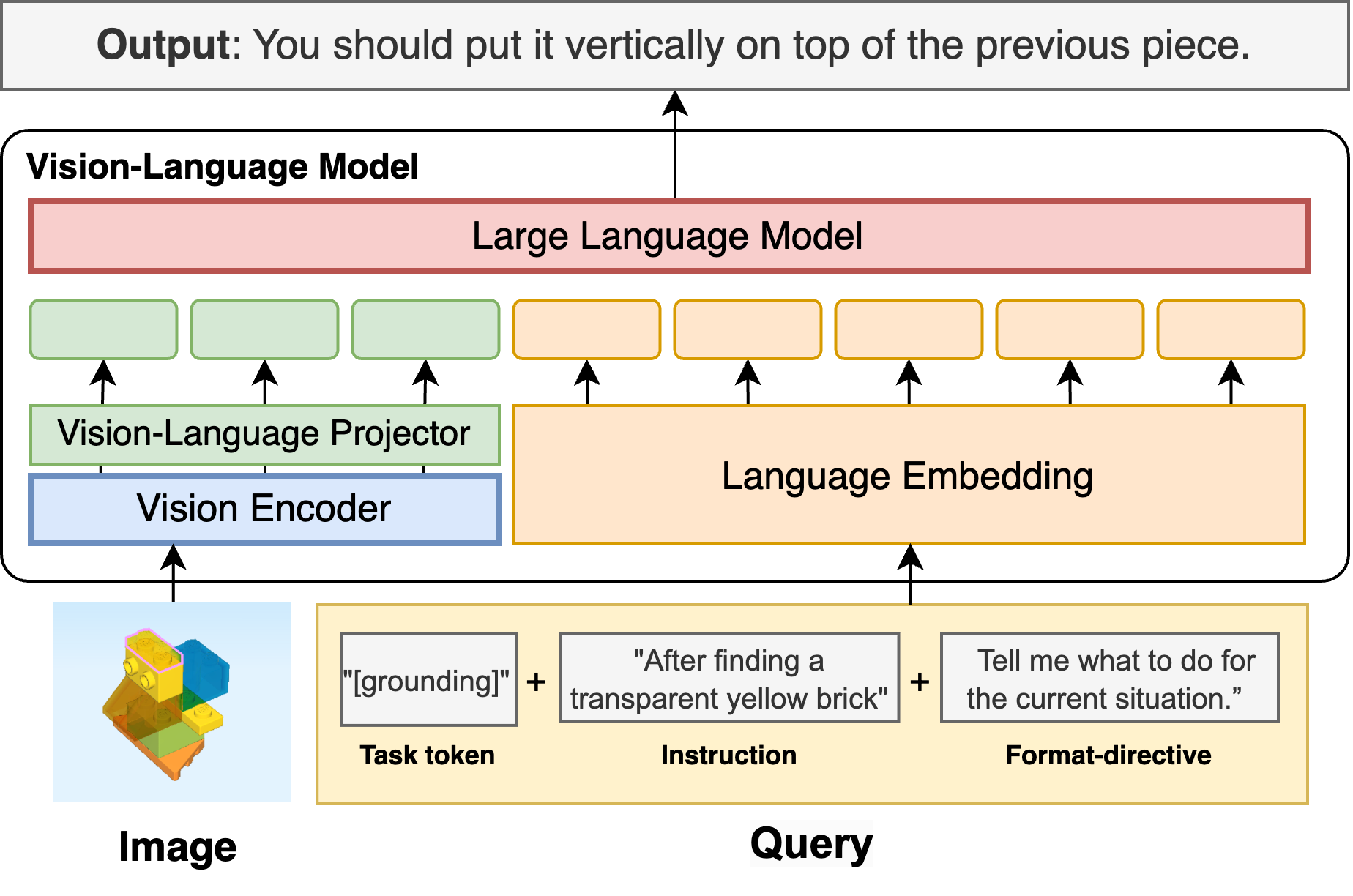} 
\caption{The architecture for one-for-all vision-language modeling. It integrates a vision encoder, a \acl{LLM}, and a vision-language projector as core functional modules, with a task-specific query for task adaptation. 
}
\label{fig:VLM}
\end{figure}

We investigated existing \acp{VLM} and derived a universal architecture for the proposed tasks, as illustrated in \autoref{fig:VLM}.
Given an image, either from manuals or real-world scenes, and a task-specific query, a \ac{VLM} based on this architecture generates a textual response as the output.
This architecture integrates a \textit{vision encoder}, a \textit{\acl{LLM}}, and a \textit{vision-language projector} as core modules, along with a task-specific \textit{query} for task adaptation.
Specifically, the vision encoder processes visual inputs, the \acl{LLM} interprets and generates textual information, and the vision-language projector aligns visual and textual data for seamless task execution.
The query can be customized using the following key components:
\begin{enumerate*}[label=(\arabic*)]
  \item \textit{Task-specific token:} 
  Special tokens such as ``[grounding]'', ``[object]'', and ``[state]'' are introduced for task T1, T2, T3, respectively, to enhance task focus and accuracy. 
  \item \textit{Instruction:} Relevant manual instructions or task directives are incorporated to provide the model with context, ensuring responses align with task requirements.
  \item \textit{Format-directive:} A directive to clarify the expected output format, ensuring outputs are precise and directly applicable.
\end{enumerate*}

\section{Dataset Creation for Task Design}\label{sec:datasest-creation}

\subsection{Manual Crawling and Scene Data Matching}
\begin{figure}[htb!]
\centering
\includegraphics[width=\columnwidth]{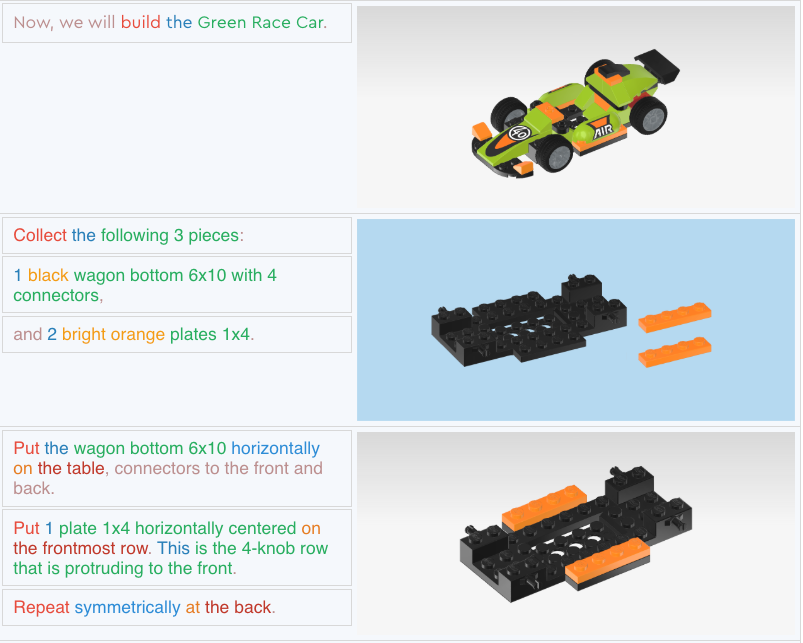}
\caption{An example of LEGO instruction manual with three sequential instruction steps. Each step includes
textual instructions paired with corresponding images
to guide the assembly process.} 
\label{fig:lego_manual} 
\end{figure}

We outline the procedure for creating the dataset for task \textbf{T1}.
First, we collected 65 official LEGO instruction manuals~\cite{lego}, designed to help blind and visually impaired users assemble LEGO sets accurately.
\autoref{fig:lego_manual} shows an example of a part of the instruction manual we crawled from the official LEGO website.
Each instruction manual comprises several elements: 
\begin{enumerate*}[label=(\arabic*)]
    \item step-by-step textual instructions;
    \item corresponding image for each step; and 
    \item tags provided by LEGO include instruction and image tags. 
\end{enumerate*}
Then, we split a full manual into several assembly sessions, each session includes two types of steps: object seeking and object assembly.
Each seeking step 
is followed by an assembly step
, forming a pair.
Next, we iterated through all sessions to generate the scene understanding dataset, $D_{T1} = \{(Q_t, V_t, A_t)\}|_{t=0}^{|D_{T1}|}$, where each element is a triplet of (query, image, text), and the query $Q_t$ is constructed by filling the query template with the task token ``[grounding]'' and the previous instruction $I_{t-1}$ as context.

\subsection{Object Position Inference}
\begin{figure}[htb!]
\centering
\includegraphics[width=\columnwidth]{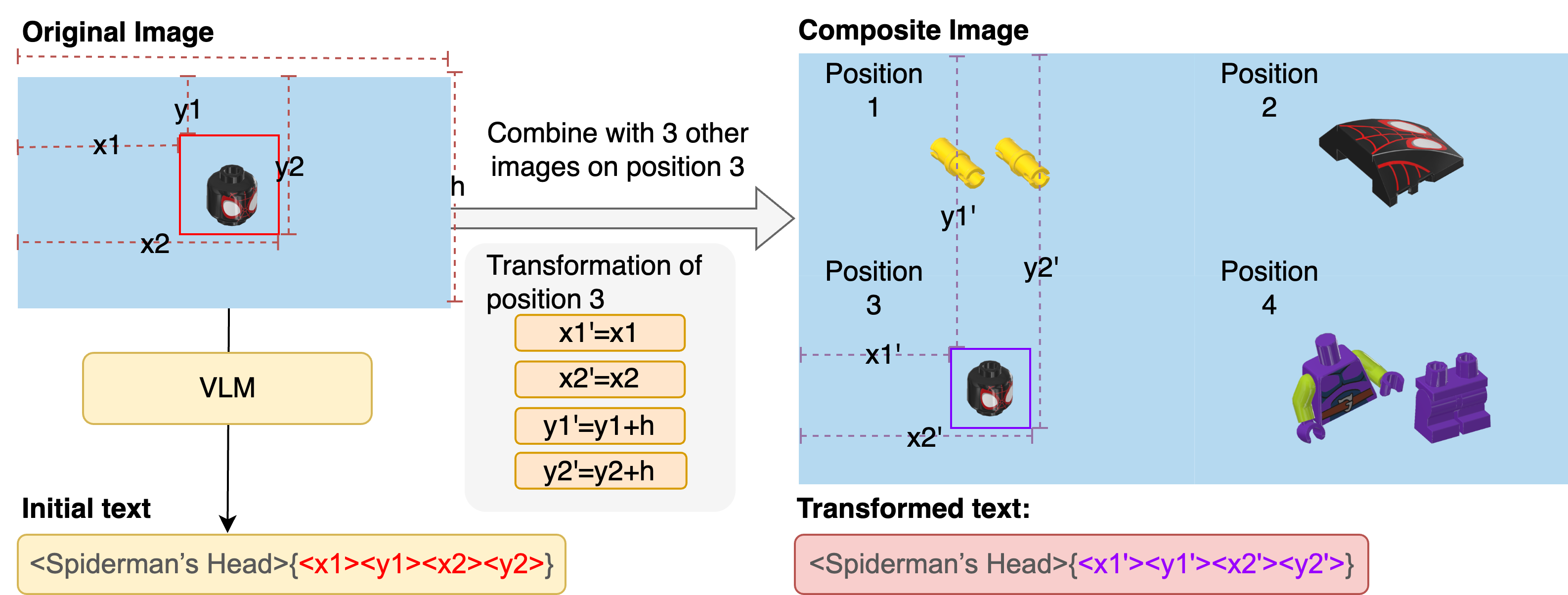}
\caption{An illustration of object detection data creation.} 
\label{fig:object_creation}
\end{figure}

This subsection outlines the procedure for creating the dataset for task \textbf{T2} as illustrated in \autoref{fig:object_creation}.
First, we iterated through all object-seeking steps, where users are asked to find specific objects.
Second, we iterated all images to generate the object detection dataset, $D_{T2} = \{(Q_{t'}, \tilde{V}_{t'}, A_{t'})\}|_{t'=0}^{|D_{T2}|}$, where each element is a triplet of (query, image, text) for a seeking step $t'=t-1$.
The query $Q_{t'}$ is constructed by filling the query template with the task token ``[object]'' and the current instruction $I_{t'}$.
The response $A_{t'}$ contains a positional text $\tilde{P}_{t'}$ formatted as ``[Object] [Xleft] [Ytop] [Xright] [Ybottom]''.
This is initially generated by querying the image using MiniGPT-v2~\citep{chen2023minigpt}, followed by rule-based filtering and manual spot checks. 
Importantly, test-set annotations are fixed and not adapted per model, preventing reward hacking or model-specific bias. 
The composite image $\tilde{V}_{t'}$ is created by combining the current image ${V}_{t'}$ with three randomly sampled images.
Last, the initial coordinates are adjusted to fit the composite image.

\subsection{Variant State Generation}
\begin{figure}[htb!]%
\centering %
\includegraphics[trim=0 10 0 0, clip, width=\columnwidth]{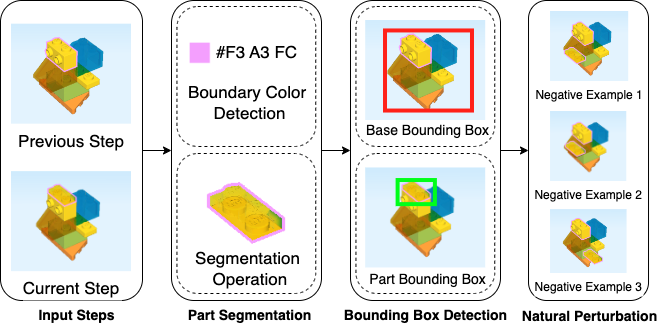} 
\caption{An illustration of synthesized negative samples for state detection.} 
\label{fig:step_creation} 
\end{figure}

This subsection outlines the procedure for creating the dataset for task \textbf{T3}, as illustrated in \autoref{fig:step_creation}.
First, we iterated through all object-assembly steps along with their corresponding previous steps, in which users are trained to assemble the parts identified in the prior steps.
Second, we conducted part segmentation by detecting the boundary colors and segmenting the parts to be assembled.
In each assembly session, the objects to be assembled in the current step are tagged in the previous step, with their boundaries highlighted in distinct colors that vary between sessions.
The highlighted colors are selected based on predominant color detection using K-means clustering~\cite{kmeans}, followed by Hue filtering~\citep{chu2014hue} and manual correction for accuracy.
Third, we added bounding boxes to the segmented objects.  
Fourth, we applied a natural perturbation to the part to be assembled by randomly shifting it within the background box, ensuring it moves by at least 5\%. 
We treated each state from the manual as a positive sample and generated three variant states as negative samples.
Then, we constructed the state detection dataset, $D_{T3} = \{(Q_{t}, V_{t}, A_{t}, \tilde{V}_{t}, \tilde{A}_{t}, \tilde{V}_{t}^{'}, \tilde{A}_{t}^{'}, \tilde{V}_{t}^{''}, \tilde{A}_{t}^{''})\}|_{t=0}^{|D_{T3}|}$, where the query $Q_t$ is constructed by filling the query template with the task token ``[state]'', the current instruction $I_{t}$, and the previous instruction $I_{t-1}$ as context.
The variant states $\{\tilde{V}_{t}, \tilde{V}_{t}^{'}, \tilde{V}_{t}^{''}\}$ are natural pertubations of $V_{t}$.
The responses $A_{t}$ and $\{\tilde{A}_{t}, \tilde{A}_{t}^{'}, \tilde{A}_{t}^{''}\}$ indicate the correct and incorrect statuses of the assembly state, respectively.
Note that perturbation variants are first generated per session, after which samples are split for train and test sets.

\subsection{Dataset Statistics and Partition}
We summarize the statistics of \OurData{} dataset in Table \ref{tab:statistic_manual}.
It is generated from 65 LEGO instruction manuals and divided into 397 sessions, covering 5,612 scenes, 4,784 objects, and 2,716 states.
It consists of 5,614 instruction steps, with each step containing an image and corresponding textual instructions. 
Out of these, 4,814 steps focus on object seeking, while 5,614 steps involve parts assembly, covering 3,172 states with 222 distinct boundary colors.
Overall, the dataset includes 35,612 vision-language data samples: 5,612 for T1, 19,136 for T2, and 10,864 for T3, respectively.
Note that all test data are strictly held out, with no overlap with training data, ensuring no data leakage.

\begin{table}[h!]
\footnotesize
\centering
\setlength{\tabcolsep}{6pt} 
\resizebox{\columnwidth}{!}{
\begin{tabular}{@{}ccccc@{}}
\toprule
\multicolumn{1}{c}{\# \textbf{Manual}} & 
\multicolumn{1}{c}{\# \textbf{Session}} & 
\multicolumn{1}{c}{\# \textbf{Scene}} & 
\multicolumn{1}{c}{\# \textbf{Object}}
& \multicolumn{1}{c}{\# \textbf{State}} \\ 
65 & 397 & 5,612  & 4,784 & 2,716 \\ \hline
\multicolumn{1}{c}{\# \textbf{Step}} & 
\multicolumn{1}{c}{Overall} & 
\multicolumn{1}{c}{Identification} & 
\multicolumn{1}{c}{Assembly} &  \\ 
 & 10,428 & 4,814 & 5,614 & \\ \hline
\multicolumn{1}{c}{\# \textbf{Sample}} & All Tasks & \multicolumn{1}{c}{T1} & \multicolumn{1}{c}{T2} & \multicolumn{1}{c}{T3} \\
Overall & 35,612 & 5,612 & 19,136 & 10,864 \\
Training & 26,709 & 4,209 & 14,352 & 8,148 \\
Testing & 8,903 & 1,403 & 4,784 & 2,716 \\
\bottomrule
\end{tabular}
}
\caption{Statistics of \OurData{} dataset.}
\label{tab:statistic_manual}
\end{table}

\section{Experimental Setup}\label{sec:experimental-setup}
\subsection{Benchmark Models}
We benchmark the following nine \acp{VLM}:
\begin{enumerate*}[nosep]
\item \textbf{mPLUG-OWL2~\citep{ye2024mplug}} replaces attention with a modality adapter in a \ac{LLM} decoder.
\item \textbf{BLIP2~\citep{li2023blip2}} uses a two-stage pre-trained Q-Former between an image encoder and a \ac{LLM} to bridge the vision-language modality gap.
\item \textbf{LLaVa~\citep{liu2023llava}} combines a visual CLIP encoder
and a language decoder Vicuna for general-purpose visual language understanding.
\item \textbf{Qwen-VL~\citep{bai2023qwen}} connects a \ac{LLM} with a visual encoder using a position-aware vision-language adapter towards fine-grained visual understanding.
\item \textbf{InstructBLIP~\citep{instructblip}} explores general-purpose vision-language instruction tuning based on the pretrained BLIP2. 
\item \textbf{MiniGPT-v2~\citep{chen2023minigpt}} links a frozen ViT visual encoder with Llama-2 via a projection layer, applicable for diverse tasks via task-specific multimodal instructions.
\item \textbf{Otter~\citep{li2023mimic}} is tuned on the OpenFlamingo, conditioning the language model on images for multi-modal perception and reasoning.
\item \textbf{MiniGPT-4~\citep{zhu2023minigpt4}} integrates a frozen visual ViT\&Q-Former encoder and Vicuna via a projection layer, unlocking advanced multimodal capabilities like GPT-4.
\item \textbf{GPT-4o}~\cite{gpt4} is a widely used commercial model accessible via APIs.
\item \textbf{Qwen-VL-2.5} update Qwen-VL with the latest Qwen \ac{LLM}~\cite{qwen2.5}.
\item \textbf{Gemini-2.5-flash}~\cite{comanici2025gemini} is a thinking model, designed to tackle increasingly complex problems.
\item \textbf{GLM-4.1-thinking}~\citep{hong2025glm} is designed to explore the upper limits of reasoning.

\end{enumerate*}

\subsection{Evaluation Metrics}
We consider multiple metrics for comprehensively evaluating specific tasks.
\textit{Scene understanding:}
\begin{enumerate*}[label=(\arabic*), nosep]
\item \textbf{F1-Theme} \jp{is the identified theme entities' F1 score that measures the harmonic mean of precision and recall.
It evaluates the accuracy of correctly mentioned theme entities in the generated instructions compared to the reference instructions, obtained from instruction manuals.}
\item \textbf{BLEU} \jp{measures precision, which measures the ratio of 1-grams in the generated responses that match those in the reference responses. 
} 
\item \textbf{ROUGE} \jp{measures recall, which calculates the ratio of 1-grams in the reference responses that are captured by the generated responses.
}
\end{enumerate*}
\textit{Object Detection:}
\begin{enumerate*}[label=(\arabic*), nosep]
\item \textbf{F1-Object} \jp{is identified object entities' F1 score. It evaluates the accuracy of identified object entities in the generated output compared to the reference data.}
\item \textbf{\Acf{IOU}} \jp{measures the overlap between the predicted and reference bounding boxes.}
\end{enumerate*}
\textit{State Detection:}
\begin{enumerate*}[label=(\arabic*), nosep]
\item \textbf{F1-State} \jp{score is a metric that combines precision and recall, measuring the model's accuracy in identifying the correct states.
It considers both precision and recall, effectively capturing the performance across both the minority (original state) and majority (generated state) classes.} 
\item \textbf{\Acf{FPR}} \jp{measures the ratio of incorrectly classified negative instances as positive, highlighting the issue of incorrect training content.}
\end{enumerate*}

\section{Outcomes}\label{sec:outcome}

\subsection{Benchmarking Results}
\label{sec:main_result}

We compare the performance of nine prevailing \acp{VLM} on the proposed \OurData{} dataset, without and with finetuing, as shown in Table~\ref{tab:benchmarks}.

\begin{table*}[htb!]
\centering
\footnotesize
\setlength\tabcolsep{6.6pt} 
\resizebox{\textwidth}{!}{
\begin{tabular}{lrrrrrrrrrrrrrr}
\toprule
Model & \multicolumn{6}{c}{T1: Scene understanding} & \multicolumn{4}{c}{T2: Object detection} & \multicolumn{4}{c}{T3: State detection} \\
\cmidrule(lr){2-7} \cmidrule(lr){8-11} \cmidrule(lr){12-15}
& \multicolumn{2}{c}{F1-Theme $\uparrow$} & \multicolumn{2}{c}{ROUGE $\uparrow$} & \multicolumn{2}{c}{BLEU $\uparrow$} & \multicolumn{2}{c}{F1-Object $\uparrow$} & \multicolumn{2}{c}{\acs{IOU} $\uparrow$} & \multicolumn{2}{c}{F1-State $\uparrow$} & \multicolumn{2}{c}{\acs{FPR} $\downarrow$} \\
 PEFT (LoRA) & w/o & w/ & w/o & w/ & w/o & w/ & w/o & w/ & w/o & w/ & w/o & w/ & w/o & w/ \\
\midrule
mPLUG-OWL2 & 23.16 & 29.70$^\blacktriangle$ & 25.23 & 32.75$^\blacktriangle$ & 3.04 & 8.46$^\blacktriangle$ & 77.94 & 93.05$^\blacktriangle$ & 14.23 & 34.88$^\blacktriangle$ &36.82& 15.00$^\triangledown$ & 63.41 & 35.00$^\blacktriangle$ \\
BLIP2 &27.85 & 32.65$^\blacktriangle$ & 29.55 & 40.35$^\blacktriangle$ & 6.40 & 12.50$^\blacktriangle$ & 77.35& 84.48$^\blacktriangle$ & \textbf{34.25} & 40.57$^\blacktriangle$ & 24.71 & 28.50$^\blacktriangle$ & 100.00 & 94.87$^\blacktriangle$ \\
LlaVA& 32.10 & 35.88$^\blacktriangle$ & 13.63 & \textbf{42.97}$^\blacktriangle$ & 0.99 & \textbf{17.73}$^\blacktriangle$ &54.39 & 72.32$^\blacktriangle$ & $\emptyset$ & \textbf{60.98}$^-$ & 25.04 & 30.00$^\blacktriangle$  & 99.41 & 50.00$^\blacktriangle$ \\
Qwen-VL & 34.84 & 37.00$^\blacktriangle$ & \textbf{29.57} & 39.08$^\blacktriangle$ & 5.19 & 13.13$^\blacktriangle$ & 78.56 & 89.15$^\blacktriangle$ & 25.60 & 30.08$^\blacktriangle$ & 39.77 & \textbf{39.53}$^\triangledown$ & 97.99 & 96.82$^\blacktriangle$ \\
InstructBLIP & 32.85 & 32.85$^-$ & 29.20 & 36.87$^\blacktriangle$ & 4.91 & 11.52$^\blacktriangle$ & 79.76& \textbf{98.16}$^\blacktriangle$ & $\emptyset$ & 47.20$^-$ & 0.00 & 0.02$^\blacktriangle$ & \textbf{1.22} & \textbf{0.00}$^\blacktriangle$ \\ 
MiniGPT-v2 & 33.06 & \textbf{37.52}$^\blacktriangle$ &34.72 & 32.56$^\triangledown$ & \textbf{9.81} & 8.28$^\triangledown$ & 84.95 & 85.91$^\blacktriangle$ & 26.98 & 25.94$^\triangledown$ & 36.76 & 38.64$^\blacktriangle$ & 60.42 & 80.72$^\triangledown$ \\ \hline
Otter & 11.49 & \multicolumn{1}{c}{/} & 12.12 & \multicolumn{1}{c}{/} & 1.28 & \multicolumn{1}{c}{/} & 72.39 & \multicolumn{1}{c}{/} & $\emptyset$ & \multicolumn{1}{c}{/} & 35.33 & \multicolumn{1}{c}{/} & 75.88 & \multicolumn{1}{c}{/} \\ 
MiniGPT-4 & 34.11 & \multicolumn{1}{c}{/} & 15.05 & \multicolumn{1}{c}{/} & 1.88 & \multicolumn{1}{c}{/} & 87.09 & \multicolumn{1}{c}{/} & 30.20 & \multicolumn{1}{c}{/} & 37.19 & \multicolumn{1}{c}{/} & 67.37 & \multicolumn{1}{c}{/} \\
GPT-4o& 25.81& \multicolumn{1}{c}{/} & 18.67 & \multicolumn{1}{c}{/} & 2.00 & \multicolumn{1}{c}{/} & 66.67 & \multicolumn{1}{c}{/} & 21.68 & \multicolumn{1}{c}{/} & \textbf{40.54} & \multicolumn{1}{c}{/} & 43.06 & \multicolumn{1}{c}{/} \\

Qwen-VL-2.5& 19.13& \multicolumn{1}{c}{/} & 15.34 & \multicolumn{1}{c}{/} & 2.70 & \multicolumn{1}{c}{/} & 75.55 & \multicolumn{1}{c}{/} & 10.63 & \multicolumn{1}{c}{/} & 33.85 & \multicolumn{1}{c}{/} & 36.70 & \multicolumn{1}{c}{/} \\

GLM-4.1-thinking&  35.73& \multicolumn{1}{c}{/} & 19.44& \multicolumn{1}{c}{/} & 2.31 & \multicolumn{1}{c}{/} & 70.64 & \multicolumn{1}{c}{/} & 4.40 & \multicolumn{1}{c}{/} & 39.61 & \multicolumn{1}{c}{/} & 38.53 & \multicolumn{1}{c}{/} \\

Gemini-2.5-flash& \textbf{37.35}& \multicolumn{1}{c}{/} & 28.28 & \multicolumn{1}{c}{/} & 6.21 & \multicolumn{1}{c}{/} & \textbf{93.12} & \multicolumn{1}{c}{/} & 13.16 & \multicolumn{1}{c}{/} & 28.81 & \multicolumn{1}{c}{/} & 8.72 & \multicolumn{1}{c}{/} \\




\bottomrule
\end{tabular}
}
\caption{\jp{Benchmarking \aclp{VLM} on \OurData{} dataset.
The upper part represents open-resourced \aclp{VLM}, both without (/wo) and with (/w) \ac{PEFT} using \ac{LoRA}. 
The lower part presents results obtained via API calls to the latest \aclp{VLM}.
The bold font indicates the highest score in each column.
Symbols $\uparrow$ and $\downarrow$ denote that higher and lower values are better, respectively.
Symbol ``-'' indicates the model is not applicable for fine-tuning. 
Symbol ``$\emptyset$'' denotes a meaningless zero as the model fails to generate output as instructed.
The superscripts ``$^\blacktriangle$'', ``$^\triangledown$'', and ``$^-$'' indicate an increase, decrease, or inapplicability in the evaluation score after fine-tuning, respectively.}}
\label{tab:benchmarks}
\end{table*}

\jp{First of all, existing models struggle with fine-grained assembly tasks in \ac{AR}.
While the F1-Object score for task T2 using the fine-tuned InstructBLIP is high at 98.16\%, the \ac{IOU} is only 47.20\%, indicating minimal overlap between the predicted object positions and the ground truth, which is still unsatisfactory.
Even the commercial model GPT-4o achieves only 66.67\% for the F1-Object score and 21.68\% on task T2.
This might also caused by the difficulty of understanding scenes.
For example, fine-tuned LlaVA shows the best overlap with reference instructions, scoring 42.97\% on ROUGE and 17.73\% on BLEU. 
However, the fine-tuned MiniGPT-v2, despite being the top performer in theme entity identification, achieves only 37.52\%.
This indicates that understanding and generating theme entities, such as LEGO parts and their properties, remains a significant challenge.
For state detection in task T3, the fine-tuned InstructBLIP achieves a 0.00\% of FPR, meaning it almost perfectly avoided incorrectly classifying negative instances as positive. 
However, the F1-State score is just 0.02\%, indicating a significant failure in identifying the correct states.
These results highlight the challenges of the proposed fine-grained vision tasks, reinforcing the need to advance this area and its resources as a valuable research topic.}

Second, open-resource \acp{VLM} with fine-tuning generally outperform or are comparable to commercial models.
For one hand, top-performing open-resource \acp{VLM} are more informative with identified accurate entities.
For example, MiniGPT-v2 with fine-tuning achieves an 11.71\% higher F1-Theme score on task T1; InstructBLIP with fine-tuning achieves a 31.48\% higher F1-Object score, while Qwen-VL shows only a 1\% decrease, compared with the commercial GPT-4o.
This may be due to the proposed dataset enhancing the model's understanding of domain-specific knowledge, such as LEGO parts and their properties.
On the other hand, top-performing open-resource \acp{VLM} can generate assembly instructions that closely align with the provided manuals.
For example, the fine-tuned LlaVA achieves ROUGE and BLEU scores that are 2.30 and 8.87 times higher than those of GPT-4o, respectively.
Besides, fine-tuning generally improves the performance of \acp{VLM}, as indicated by the results marked with the superscript ``$^\blacktriangle$''.
This highlights the significant difference between the proposed dataset and those used to train general commercial models.

Last, task difficulty varies significantly, with the biggest challenge being the alignment of positional information in images with the corresponding textual information from the query.
Specifically, in comparison to task T1, tasks T2 and T3 present greater challenges in instruction following.
For example, several \acp{VLM} (i.e., LlaVA, InstructBLIP, Otter), without fine-tuning, fail to follow instructions to generate positional information for evaluating \ac{IOU}, indicated by the symbol ``$\emptyset$.''
Additionally, all evaluated \acp{VLM} have F1-State scores below 50\%, indicating that their predictions are even worse than random guessing.
One potential reason task T3 is complex, and relies on task T2 is that it requires positional information to detect states, as well as insights from task T1 to understand the LEGO components and their relationships in the scene. 
We will explore this in future work, as this study primarily focuses on proposing the tasks rather than complex modeling.

\subsection{Data Quality Assessment}
To ensure the quality of the generated data for tasks T2 and T3, we sampled 100 data samples and conducted quality assessments for each task.
We added bounding boxes based on the coordinates and asked three annotators to evaluate the generated data quality, focusing on coordinates, entities, and negative samples, based on the following criteria:
\begin{enumerate*}[label=(\arabic*), nosep]
    \item \textbf{Entity disambiguity} measures how clearly a target entity is relevant to the scene in T2, with scores of 0, 1, and 2 indicating low, medium, and high disambiguation, respectively.
    \item \textbf{Boundary precision} measures how accurately a bounding box encloses a target object in T2, with scores of 0, 1, and 2 indicating low, medium, and high precision, respectively.
    \item \textbf{State relevance} measures whether the generated parts in an image are a relevant variation of the original image in T3.
    \item \textbf{State identifiability} measures whether the generated parts in an image are a recognizable variation of the original image in T3.
\end{enumerate*}

\begin{figure}[ht]
\centering 
\includegraphics[width=0.95\columnwidth]{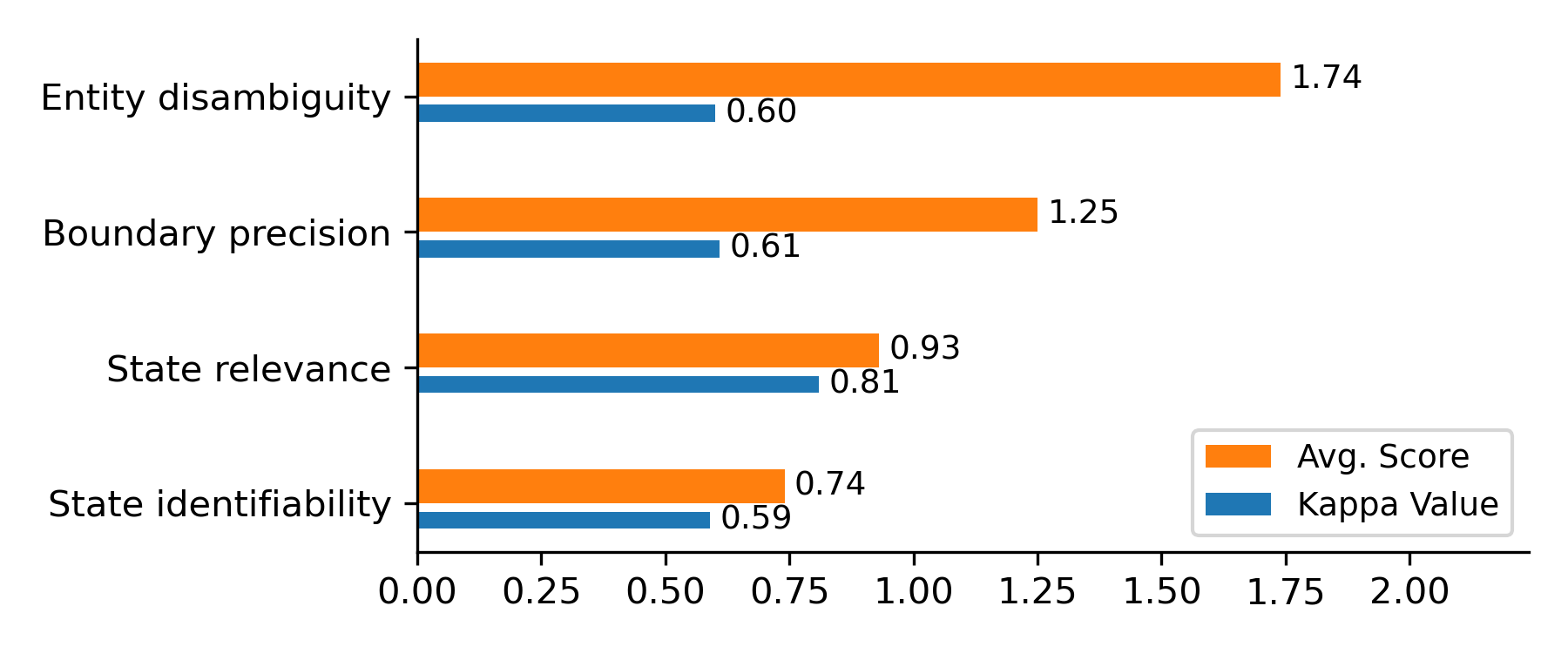} 
\caption{Human assessment of data quality.}\label{fig:data_quality}
\end{figure}

As shown in \autoref{fig:data_quality}, the average entity disambiguation and boundary precision scores are 1.74 and 1.25 out of 2  with the Kappa values~\citep{kappa} of 0.60 and 0.61, respectively.

This suggests that annotators reached a moderate agreement on the clarity of the evaluated objects in relation to the corresponding scenes.
The average state relevance and identifiability scores are 0.93 and 0.74 out of 1.00, respectively, with the Kappa scores 0.81 and 0.59. 
This indicates that annotators reached near-perfect agreement on the relevance of the generated states to the original while showing moderate agreement on the recognizable variation of those states. 
Besides, the first five data points mentioned in the guidelines were used as a sanity check to ensure that the annotators correctly understood and applied the evaluation criteria.
This ensures the quality of the generated states.

\section{Analysis}
\subsection{Impact of Query Construction}
We explore the impact of query construction on the QWEN-VL model in the early phase to identify the optimal configuration for achieving the best inference results. 

To this end, we conducted experiments on various query settings to determine which components most significantly influence performance. 
The model was chosen for this study due to its strong results under our base setting, which included both instruction directives and in-context examples.

\begin{table}[h!]
\centering
\scriptsize
\setlength{\tabcolsep}{0.5pt} 
\resizebox{\columnwidth}{!}{
\begin{tabular}{ccccccccccc}
\toprule
\multicolumn{3}{c}{Setting}& \multicolumn{3}{c}{T1} & \multicolumn{2}{c}{T2} & \multicolumn{2}{c}{T3} \\
\cmidrule(lr){1-3} \cmidrule(lr){4-6} \cmidrule(lr){7-8} \cmidrule(lr){9-10}
& Instruction & Example & T\_ACC $\uparrow$ & ROUGE $\uparrow$ & BLEU $\uparrow$ & E\_ACC $\uparrow$ & IOU $\uparrow$ & F1-State $\uparrow$ & FPR $\downarrow$ \\
\midrule
  &\Checkmark  & \Checkmark  & \textbf{11.90} & \textbf{29.57} & \textbf{5.19} & \textbf{32.41} & \textbf{25.60} & \textbf{39.77} & 97.99\\
  & \XSolidBrush & \Checkmark  & 6.84 & 24.63 & 4.58 & 5.56 & 22.01 & 34.00 & 97.50 \\
  &\Checkmark  & \XSolidBrush & 10.43 & 23.00 & 4.24 & 26.37 & 15.55 & 34.50 & 97.20 \\
  &\XSolidBrush   &\XSolidBrush & 5.27 & 11.02 & 1.30 & 3.70 & 0.10 & 25.50 & \textbf{98.40} \\
\bottomrule
\end{tabular}
}
\caption{A study of query construction results across various tasks and metrics. 
The symbol $\uparrow$ indicates higher is better, $\downarrow$ indicates lower is better, and values in bold represent the highest results.}
\label{table:query_study}
\end{table}

As shown in Table~\ref{table:query_study}, our findings highlight the critical role of query design across different tasks. The results indicate that providing both instructions and examples yields the best overall performance, demonstrating their complementary roles in guiding inference.
In the scene understanding task (T1), the presence of instructions plays a critical role in improving accuracy. When instructions are included, theme accuracy (T\_ACC) increases significantly, suggesting that clear guidance helps the model better understand scene context and thematic elements. Furthermore, both ROUGE and BLEU scores, which evaluate the structure and coherence of generated outputs, improve when at least one of the two components (instructions or examples) is present. The lowest performance in this task occurs when neither component is provided, confirming that structured input is essential for meaningful inference.

In the object detection task (T2), the results suggest that examples contribute more to entity recognition and output formatting than instructions alone. The entity accuracy (E\_ACC) drops sharply from 32.41\% to 5.56\% when examples are removed, even when instructions remain. Similarly, IoU, which measures how well the output adheres to the expected format, significantly decreases in the absence of examples. This highlights that examples provide concrete reference points, enabling the model to generate well-structured and accurate entity labels.
In the state detection task (T3), despite improvements in other metrics with enhanced query settings, the false positive rate (FPR) remains consistently high (above 97\%) across all conditions. This suggests that the model has a strong affirmative bias, likely defaulting to generating positive responses regardless of query refinements. While instructions and examples improve structured outputs, they do not sufficiently mitigate the issue of excessive false positives.

These findings emphasize the importance of query design in optimizing model performance. Combining instructions with instruction directives and in-context examples is crucial for achieving high accuracy, structured outputs, and reliable entity recognition. 
However, the persistent issue of high FPR suggests the need for further refinements, such as introducing negative examples, confidence calibration, or additional fine-tuning strategies to reduce overconfidence in affirmative responses. Future work should explore these approaches to enhance the model’s robustness and reliability in diverse inference tasks.

\subsection{Case Study of Fine-grained \ac{VLM} Challenges}\label{sec:case_study}

\begin{table*}[t!]
    \centering
    \footnotesize
    \begin{tabular}{p{14cm} p{3cm}}
        \toprule
        \textbf{Query}: [\textit{grounding}] After I find \textit{1 transparent mailbox front 2x2}. Tell me what to do for the current situation. &
        \multirow{2}{*}{\includegraphics[width=1.5cm, height=1cm]{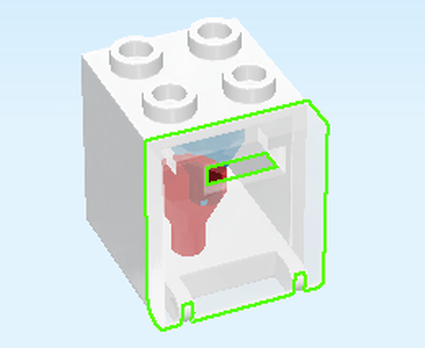}} \\
        \\
        \textbf{Reference of Assembly Instruction:} \\
        Insert the two tabs at the bottom of the mailbox casing into the two small notches on the mailbox front.  \\ 
        \hline
        \multicolumn{2}{l}{
        \textbf{mPLUG-OWL2:} Place the \textit{mailbox front 2x2} next to the block on the table}. \\
        \hline
        \multicolumn{2}{l}{
        \textbf{BLIP2:} Put the \textit{mailbox front 2x2} horizontally on the table, clasp to the back. Put the mailbox back 2x2 vertically on top, slope to the left.} \\
        \hline
        \multicolumn{2}{p{17cm}}{\textbf{LlaVA:} In the image, there is a small Lego-like object with a red liquid inside of it. The object is placed in a blue sky, and it appears to be a 3D model. To create a transparent mailbox, you need to find a \textit{transparent mailbox front that is 2x2} in size. Once you find it, you can use it as the base for your transparent mailbox. You can then place the small Lego-like object inside the mailbox to complete the scene.} \\
        \hline
        \multicolumn{2}{l}{\textbf{Qwen-VL:} Put the \textit{transparent mailbox front 2x2} on the right side of the white arch.} \\
        \hline
        \multicolumn{2}{l}{\textbf{InstructBLIP:} Place it horizontally on the table, clasp to the back.} \\
        \hline
        \multicolumn{2}{p{17cm}}{\textbf{MiniGPT-v2:} Here's what you need to do with the pieces you have collected:
        1. Place the \textit{transparent mailbox front 2x2} on top of the plate 1x2, aligned with the short side of the plate. Now you have completed the current situation. Let me know if you need help with anything else!} \\
        \hline
        \multicolumn{2}{l}{\textbf{Otter:} In view of the reasoning `A mailbox is small and square with a see through window.', the optimal answer should be: build structure.} \\
        \hline
        \multicolumn{2}{l}{\textbf{MiniGPT-4:} 
        Attach the \textit{transparent mailbox front} to the base structure, aligning the tabs with the notches.} \\
        \hline
        \multicolumn{2}{l}{\textbf{GPT-4o:} Ensure the \textit{transparent mailbox front 2x2} is securely attached by inserting the small shaft into the round brick 1x1 before proceeding with the next step.} \\
        \bottomrule
    \end{tabular}
    \caption{An example for understanding \ac{VLM} challenges. 
    }
    \label{tab:scene_ep}
\end{table*}

In this section, we present illustrative examples to highlight the capabilities and challenges in fine-grained vision-language understanding, particularly in achieving overlap with the reference and accurately recognizing entities and their properties, such as size, color, and etc.

Table~\ref{tab:scene_ep} presents an example showcasing the intuitive results of the prevailing \acp{VLM} on scene understanding.
Compared with the reference, existing \ac{VLM} models face key challenges, including: 
\begin{enumerate*}[nosep, label=(\arabic*)]
    \item \textit{Vague and generic assembly instructions.} 
        The mPLUG-OWL2 and BLIP2 models produce broad, non-specific placement directives (e.g., ``Place the mailbox front 2x2 next to the block on the table'' or ``Put the mailbox front 2x2 horizontally on the table, clasp to the back''), falling short of the precise, step-by-step guidance found in the reference. 
        The Otter's response is even more superficial, merely stating ``build structure'' without actionable detail.
    \item \textit{Struggles with fine-grained entity recognition and attribute grounding.} 
        While models like Qwen-VL and MiniGPT-v2 correctly mention the ``transparent mailbox front 2x2,'' other models such as InstructBLIP, overlook critical attributes (e.g., referring only to ``it'' or ``the mailbox front''). 
        The LlaVA model, despite its verbosity, hallucinates details (e.g., ``red liquid inside'' and ``blue sky'') absent from the scene, highlighting confusion in entity identification and property association.
    \item \textit{Tendency to hallucinate or misinterpret assembly steps.} 
        The MiniGPT-v2 and GPT-4o models introduce extraneous or incorrect actions, such as ``aligning with the short side of the plate'' or ``inserting the small shaft into the round brick 1x1,'' which are not part of the actual assembly process. This reveals a shallow grasp of both scene context and the logical flow of assembly operations.
\end{enumerate*}
Overall, while some models can partially identify the correct entities and occasionally mention relevant properties, none consistently generate instructions that are both factually accurate and contextually appropriate for fine-grained assembly tasks.
To address this gap, we benchmark the \ac{VLM} tasks and provide comprehensive resources to advance research in this field.

\subsection{Non-VLM Baseline Comparison}
Regarding the object detection task (T2), classical perception-only models (e.g., the YOLO series) achieve strong performance on predefined, fixed label sets (IoU 0.62–0.73 on 50 classes). However, they are less suitable for our setting, where the space of descriptive labels is large and each class has limited training examples. In contrast, our benchmark model achieves comparable performance (IoU 0.61) while supporting descriptive object outputs without restricting the label space (see an example in \autoref{tab:complex_example}). Besides, our setting requires models to jointly handle all tasks, integrating perception, reasoning, and the generation of spatial information in a language-compatible format, thereby encouraging the training of a unified architecture.

\begin{table}[h!]
\scriptsize
\centering
\setlength{\tabcolsep}{2pt} 
\resizebox{\columnwidth}{!}{
\begin{tabular}{ll}
\toprule
\textbf{Query}  & \begin{tabular}[c]{@{}l@{}}Please identify and highlight the positions of the following 2 pieces: 1 olive green pair \\ of legs, and 1 dark green, mossy green and bright yellowish green upper part of the body.\end{tabular}\\
\midrule
\textbf{Output} & \begin{tabular}[c]{@{}l@{}}\textless{}p\textgreater{}1 olive green pair of legs\textless{}/p\textgreater  \{\textless{}13\textgreater{}\textless{}17\textgreater{}\textless{}24\textgreater{}\textless{}37\textgreater{}\}\\ \textless{}p\textgreater{}1 dark green, mossy green and bright yellowish green upper part of the body\textless{}/p\textgreater \\ \{\textless{}24\textgreater{}\textless{}17\textgreater{}\textless{}39\textgreater{}\textless{}37\textgreater{}\}\end{tabular}
\\ \bottomrule
\end{tabular}
}
\caption{An example of language-compatible object detection. }
\label{tab:complex_example}
\end{table}

\subsection{Fine-tuning Performance Degradation} 
While fine-tuning improves performance in most settings, we observe degradation in some cases, such as Qwen-VL on T3 and MiniGPT-v2 on T2/T3 (see \autoref{tab:benchmarks}). 
We attribute this to \textit{over-specialization to surface patterns} and \textit{catastrophic forgetting} in models not designed for precise spatial supervision. 
LEGO tasks require pixel-level alignment and relational consistency, which can conflict with language-dominant pretraining objectives. 
Similar effects have been reported in prior VLM fine-tuning studies~\cite{zhai2024investigating}.

\section{Related Work}\label{sec:realated-work}
\subsection{Vision-Language Datasets for Assembly}
Various datasets have been developed to enhance assembly tasks by supporting key capabilities T1, T2, and T3. A comparative summary is presented in Table~\ref{tab:comparision}.

\begin{table}[ht!]
\footnotesize
\centering
\setlength{\tabcolsep}{1.8em}
\resizebox{\columnwidth}{!}{
\begin{tabular}{lcccr}
\toprule
\textbf{Dataset} & \textbf{T1} & \textbf{T2} & \textbf{T3} & \textbf{Size} \\
\toprule
COIN & \centering \Checkmark & \centering \XSolidBrush & \centering \Checkmark & 11,827 \\
HowTo100M & \centering \Checkmark & \centering \XSolidBrush & \centering \Checkmark & 23,611 \\
TEACh & \centering \Checkmark & \centering \XSolidBrush & \centering \Checkmark & 3,215 \\
MIMIC-IT & \centering \Checkmark & \centering \XSolidBrush & \centering \XSolidBrush & 2.8M \\
HoloAssist& \centering \Checkmark & \centering \XSolidBrush & \centering \Checkmark & 350 \\
EPIC-KITCHENS & \centering \XSolidBrush & \centering \Checkmark & \centering \Checkmark & 89,977 \\
Assembly101 & \centering \Checkmark & \centering \XSolidBrush & \centering \Checkmark & 4,321 \\
Cross-task & \centering \XSolidBrush & \centering \Checkmark & \centering \Checkmark & 4,713 \\
RareAct & \centering \XSolidBrush & \centering \Checkmark & \centering \Checkmark & 7,607 \\
\hline
\OurData{} (Ours) & \centering \Checkmark & \centering \Checkmark & \centering \Checkmark & 35,612 \\
\bottomrule
\end{tabular}
}

\caption{\jp{Compariable datasets for assembly tasks towards \ac{AR} training regarding the cabablities of  \textbf{T1}, \textbf{T2}, \textbf{T3}.
The symbols \Checkmark and \XSolidBrush indicate the presence or absence of each capability.}}
\label{tab:comparision}
\end{table}

COIN provides a strong foundation for sequential task analysis with comprehensive annotations, aiding research in multimodal learning \citep{tang2019coin}. HoloAssist captures egocentric human-AI interactions using mixed-reality headsets, offering valuable real-world insights \citep{wang2023holoassist}. HowTo100M, with its extensive collection of 136 million video clips and transcribed narrations, is highly effective for text-to-video retrieval but lacks fine-grained object annotations \citep{miech2019howto100m}. TEACh focuses on interactive dialogues in domestic environments, improving dialogue modeling but not prioritizing detailed object detection \citep{padmakumar2021teach}.

While some datasets excel in action recognition, they often lack the procedural depth necessary for effective training. Assembly101, with over 4,000 toy assembly videos, does not include step-by-step instructional details \citep{sener2022assembly101}. CrossTask facilitates weakly supervised learning by leveraging narrations and step lists but lacks temporal annotations to clarify action sequences \citep{zhukov2019crosstask}. EPIC-KITCHENS provides extensive annotations on unscripted kitchen activities, yet it does not offer structured procedural guidance \citep{Damen2018EPICKITCHENS}. RareAct presents unique interactions, challenging models to interpret complex actions without explicit instructions \citep{miech20rareact}.
\citet{stanescu2023state} introduce a state-aware prior that significantly improves object detection in assembly tasks like furniture and Lego construction. However, it lacks textual descriptions of object states and does not address fine-grained detection \citep{stanescu23stateaware}. Meanwhile, MIMIC-IT contributes 2.8 million multimodal instruction-response pairs, enriching conversational modeling but lacking scenario-specific descriptions that would enhance task guidance \citep{li2023mimic}.

To bridge these gaps, we conduct a comparative analysis evaluating dataset coverage of essential components (objects, states, and scenes) critical for developing AR-guided assistants. Additionally, we introduce a new simulated dataset featuring nearly 400 objects, systematically alternating between object detection and assembly functions. This dataset is designed to optimize the training and evaluation of advanced \acp{VLM}, improving their ability to handle fine-grained assembly tasks.

\subsection{Fine-grained Vision-Language Models}
Recent advancements in vision-language models (VLMs) have broadened their applications, enhancing accessibility, search indexing, and interactive content retrieval \citep{fanpoly,zhang2024vision,du2022survey,gan2022vision}. These models generate captions, describe images \citep{zhou2020unified,hu2022scaling}, facilitate visual QA \citep{bazi2023vision}. For instance, they can accurately identify objects in images based on phrases like ``the red car'' \citep{subramanian2022reclip}. Furthermore, these models facilitate conversational interactions about visual elements \citep{chen2022unsupervised}, which makes virtual assistants more adept at handling inquiries related to images and videos, which makes virtual assistants more adept at handling inquiries related to images and videos. Additionally, VLMs can support multilingual content \citep{gwinnup2023survey}, benefiting assistive technologies and education \citep{chi2020just,wang2024lave}.

Despite progress, VLMs often lack fine-grained modeling for scene understanding, object recognition, and error detection. While CLIP demonstrates strong generalization \citep{radford2021learning}, models like EfficientVLM \citep{wang2023efficientvlm}, MiniGPT-v2 \citep{chen2023minigpt}, and Qwen-VL \citep{bai2023qwen} struggle with intricate tasks. OSCAR \citep{li2020oscar} and VisionLLM \citep{wang2024visionllm} improve image-text alignment but face challenges in context-specific adaptation. The Otter model advances sequential task management yet falls short in detailed analysis \citep{li2023otter}.

Efforts in procedural video representation \citep{zhong2023learning} and instructional task graphs \citep{ashutosh2023videomined} highlight potential VLM integration but reveal gaps in error correction. Hence, we propose a vision-language architecture optimized for fine-grained tasks, advancing training and benchmarking of VLMs.

\section{Conclusion}\label{sec:conclusion}
Vision-language models have achieved significant success in general vision-language tasks, but they often struggle with fine-grained understanding, particularly in structured instructional scenarios.
This research explores fine-grained vision-language modeling for manual-guided LEGO assembly tasks, focusing on scene understanding, object detection, and state detection.
We created a specialized multimodal dataset from LEGO instruction manuals, designing fine-grained tasks to evaluate VLM performance in tracking assembly sequences and interpreting instructions. 
Using a unified architecture, we assessed several models and found that they struggle with fine-grained scene understanding and assembly state detection, highlighting the need for further benchmarking and improvement.
Beyond technical advancements, our research aims to empower blind and visually impaired individuals by enabling AI-driven learning tools, promoting greater accessibility and independence.

\section*{Safe and Responsible Innovation Statement}\label{sec:discussion}
\subhead{Implications.}
The implications span beyond research: such systems could democratize hands-on learning by enabling visually impaired users to access assembly tasks through multimodal feedback, and could revolutionize technical education and vocational training by offering consistent, language-grounded support. Our findings further highlight the current limitations of \acp{VLM} in precise reasoning and verification, pointing toward future directions in responsible AI development that benefit society at large—including learners, workers, and underserved communities.
The architecture is intentionally lightweight, serving as a unified evaluation wrapper that enables fair comparison across diverse VLMs without introducing confounding model-specific designs. Our motivation is not to propose a new architecture, but to expose capability gaps in existing multimodal assistants under a consistent interface.

\subhead{Limitations.} 
We also acknowledge the limitations, including:
First, simulated evaluation scenarios.
While T1 is constructed from real-world data, T2 and T3 are generated using rule-based programs, rather than diverse real-world assembly scenarios.
Moreover, we only consider 2D settings rather than 3D environments, which represents a major gap between our benchmarks and real-world scenarios.
Second, annotation bias. 
In T2, we only use model-assisted annotation to create a scalable benchmark dataset without full-scale human correction.
Although we sampled 100 data instances for human assessment of data quality, human correction was not performed on the entire dataset.
Last, fidelity of evaluation metrics.
While BLEU and ROUGE provide complementary measures of lexical overlap and semantic alignment, they do not fully capture procedural correctness or the functional equivalence of assembly instructions.

\subhead{Future work.} 
We plan to extend this benchmark towards more realistic scenarios, including the construction of datasets collected from users interactions with LEGO assembly assistants in XR environments. 
We also plan to develop more comprehensive evaluation metrics that better reflect task-level performance, including procedural accuracy, functional validity, and user-centered effectiveness.

\subhead{Ethical Considerations.} We recognize the ethical implications of developing \acp{VLM} for user interactions, so addressing these concerns is essential. To ensure ethical standards, we benchmark open-source \acp{VLM} and prioritize user-centric design, such as inclusiveness for diverse users.


\begin{acks}
We would like to thank all reviewers for their time, effort, and constructive comments on our paper. 
We appreciate their feedback and have addressed their concerns in the camera-ready version.
\end{acks}

\bibliographystyle{ACM-Reference-Format}
\bibliography{icmi26}

\appendix

\section{System Implementation Details}
During data synthesis, we utilized MiniGPT4-v2~\citep{chen2023minigpt} as the \ac{VLM}, employing straightforward rules and human correction to create vision-language data.
To evaluate the performance of benchmark models, we randomly sampled 25\% of the dataset to serve as the test set. 
All inference and fine-tuning experiments were conducted on NVIDIA A100 SXM4 40GB GPUs.
We set the learning rate to 1e-5 and the batch size to 16 for our experiments. 
The AdamW optimizer~\citep{loshchilov2017decoupled} was employed to facilitate efficient training, which spanned 3 epochs.

\section{Query Templates}
We investigate three fine-grained vision-language tasks and have devised specific query templates for each, as shown in Table \ref{table:prompt_template}. 
\begin{table}[h]
\centering
\scriptsize 
\begin{tabular}{@{}p{\columnwidth}@{}} 
\toprule
\textbf{T1:} Scene Understanding \\
\textbf{Token:} [grounding] \\
\textbf{Selection Rule:} [step\_symbol] with [step]. Instruction from this step. \\
\textbf{Format-Directive:} Tell me what to do for the current situation. \\
\hline
\textbf{T2:} Object Detection \\
\textbf{Token:} [object] \\
\textbf{Selection Rule:} [step\_symbol] with [eop], Word starts with ``find'' or ``collect''. Instruction from this step. \\
\textbf{Format-Directive:} Providing the positions in the format: \{[object] [Xleft][Ytop][Xright][Ybottom]\}, 
with X and Y coordinates normalized to [0,100]. [Xleft] and [Ytop] for the top-left corner. [Xright] and [Ybottom] for the bottom-right corner. \\
\hline
\textbf{T3:} State Detection \\
\textbf{Token:} [state] \\
\textbf{Selection Rule:} [step\_symbol] with [step]. Previous [step\_symbol] is [eop]. Instruction from this step and the previous step. \\
\textbf{Format-Directive:} Just tell me Yes or No. \\
\bottomrule
\end{tabular}

\caption{Query Templates for Fine-Grained \ac{VLM} Tasks.}
\label{table:prompt_template}
\end{table}

\begin{table}[htb!]
\centering
\scriptsize
\begin{tabular}{|p{0.95\columnwidth}|} 
\hline
\texttt{ \{ } \\
\texttt{ \ \ "instruction\_id": 95,} \\
\texttt{ \ \ "text": [ "Find 2 transparent red flat tiles 1x2." ],} \\
\texttt{ \ \ "VLM": \{ } \\
\texttt{ \ \ \ \ "img\_path": "...",} \\
\texttt{ \ \ \ \ "parts\_img\_path": "None",} \\
\texttt{ \ \ \ \ "id\_sign": "Car",} \\
\texttt{ \ \ \ \ "step\_num": 28,} \\
\texttt{ \ \ \ \ "step\_class": "eop",} \\
\texttt{ \ \ \ \ "other\_sign": "AG",} \\
\texttt{ \ \ \ \ "background\_rect": "None",} \\
\texttt{ \ \ \ \ "color\_list": [],} \\
\texttt{ \ \ \ \ "bound\_color": [],} \\
\texttt{ \ \ \ \ "task\_label": "object",} \\
\texttt{ \ \ \ \ "query": "Find 2 transparent red flat tiles 1x2.",} \\
\texttt{ \ \ \ \ "MiniGPTv2\_output": "<p>2 transparent red flat tiles</p> \{<28><17><73><50>\}" } \\
\texttt{ \ \ \}, } \\
\texttt{ \ \ "img": "...",} \\
\texttt{ \ \ "entities": [ } \\
\texttt{ \ \ \ \ \{ "this\_line": "Find 2 transparent red flat tiles 1x2." \},} \\
\texttt{ \ \ \ \ \{ "verb": "Find" \},} \\
\texttt{ \ \ \ \ \{ "det": "2" \},} \\
\texttt{ \ \ \ \ \{ "colorinfo": "transparent red" \},} \\
\texttt{ \ \ \ \ \{ "theme": "flat tiles 1x2" \},} \\
\texttt{ \ \ \ \ \{ "suffix": "." \} ] } \\
\texttt{ \} } \\
\hline
\end{tabular}
\caption{An example of a JSON schema for integrated multimodal data.}
\label{tab:json_VLM_config}
\end{table}

\section{Data Schema}\label{app:ex_json}
This section presents the JSON data schema for integrating multimodal data (visual and textual) and fine-tuning \ac{VLM}. The schema defines key components that structure and organize data efficiently, ensuring seamless preprocessing before final compilation.
The key fields related to \ac{VLM} in the JSON schema include:
\begin{itemize}
    \item \textbf{img\_path}: Stores the local path of images downloaded via web scraping, pointing to the original files fetched from the web.
    \item \textbf{id\_sign}: Extracted from the image filename, this field identifies the sub-assembly session (e.g., ``CAR'' in this example).
    \item \textbf{step\_num}: Represents the sequence number of the assembly step (e.g., step 28), tracking the assembly's progression.
    \item \textbf{step\_class}: A classification tag (e.g., ``eop'' for ``end of part'') indicating an object-finding step in the assembly process.
    \item \textbf{MiniGPTv2\_output}: Stores processed output from MiniGPTv2, containing annotations or modifications essential for subsequent dataset creation.
    \item \textbf{parts\_img\_path}, \textbf{bound\_color}, \textbf{color\_list}: Reserved for assembly steps classified as ``step'' and crucial for state detection dataset generation:
    \begin{enumerate*}
        \item \textbf{color\_list}: Records all possible boundary colors used throughout the assembly session.
        \item \textbf{bound\_color}: Consolidates the most probable boundary colors through a voting mechanism, ensuring consistency across the dataset.
    \end{enumerate*}
\end{itemize}

To support fine-tuning, our default JSON format follows the Llava model's fine-tuning data structure, ensuring an optimized and adaptable framework for general use cases.
An example of the JSON schema for VLM fine-tuning is illustrated in Table~\ref{tab:json_example_processed}.
\begin{table}[htb!]
\centering
\footnotesize
\begin{tabular}{|p{0.95\columnwidth}|} 
\hline
\texttt{ \{ } \\
\texttt{ \ \ "id": "lego-60263-ocean-mini-submarine-readscr-12",} \\
\texttt{ \ \ "image": "sm01-island-0\_step\_Step\%20\%23\%23E.png",} \\
\texttt{ \ \ "conversations": [} \\
\texttt{ \ \ \ \ \{ } \\
\texttt{ \ \ \ \ \ \ "from": "human",} \\
\texttt{ \ \ \ \ \ \ "value": "[grounding] After I find 1 reddish brown chest bottom 2x4.} \\
\texttt{ \ \ \ \ \ \ Tell me what to do for the current situation."} \\
\texttt{ \ \ \ \ \},} \\
\texttt{ \ \ \ \ \{ } \\
\texttt{ \ \ \ \ \ \ "from": "gpt",} \\
\texttt{ \ \ \ \ \ \ "value": "Put it horizontally on the table."} \\
\texttt{ \ \ \ \ \} ] } \\
\texttt{ \} } \\
\hline
\end{tabular}
\caption{An example of a JSON schema for \ac{VLM} fine-tuning.}
\label{tab:json_example_processed}
\end{table}
The full dataset, comprising JSON and PNG files, occupies approximately 11.8 GB. 
For specialized requirements, such as those for the Otter model, we convert all text and image data into Parquet files.

\section{Case Study of Object Detection}
Table~\ref{tab:obj_ep} present an example of object detection task. 
We obtain the following observations:
First, only InstructiBLIP and Otter generate the full description of the object ``1 light purple brick 2x6'' without disambiguation.
Second, none of the models accurately predict the precise positions as outlined in the reference.
\begin{table}[htb!]
    \centering
    \footnotesize
    \begin{tabular}{p{1.4cm} p{4cm} p{2cm}}
        \toprule[1.1pt]
         & \textbf{Text} & \textbf{Image} \\
        \midrule
        \textbf{Query} & 
        [object] Please identify and highlight the positions of 1 light purple brick 2x6. Providing the positions in the format: object {[Xleft,Ytop,Xright,Ybottom]}, with X and Y coordinates normalized to [0,100]. Xleft and Ytop for the top-left corner. Xright and Ybottom for the bottom-right corner. & 
        \multirow{6}{*}{\includegraphics[width=2.2cm, keepaspectratio, clip, trim=0 1cm 0 1cm]{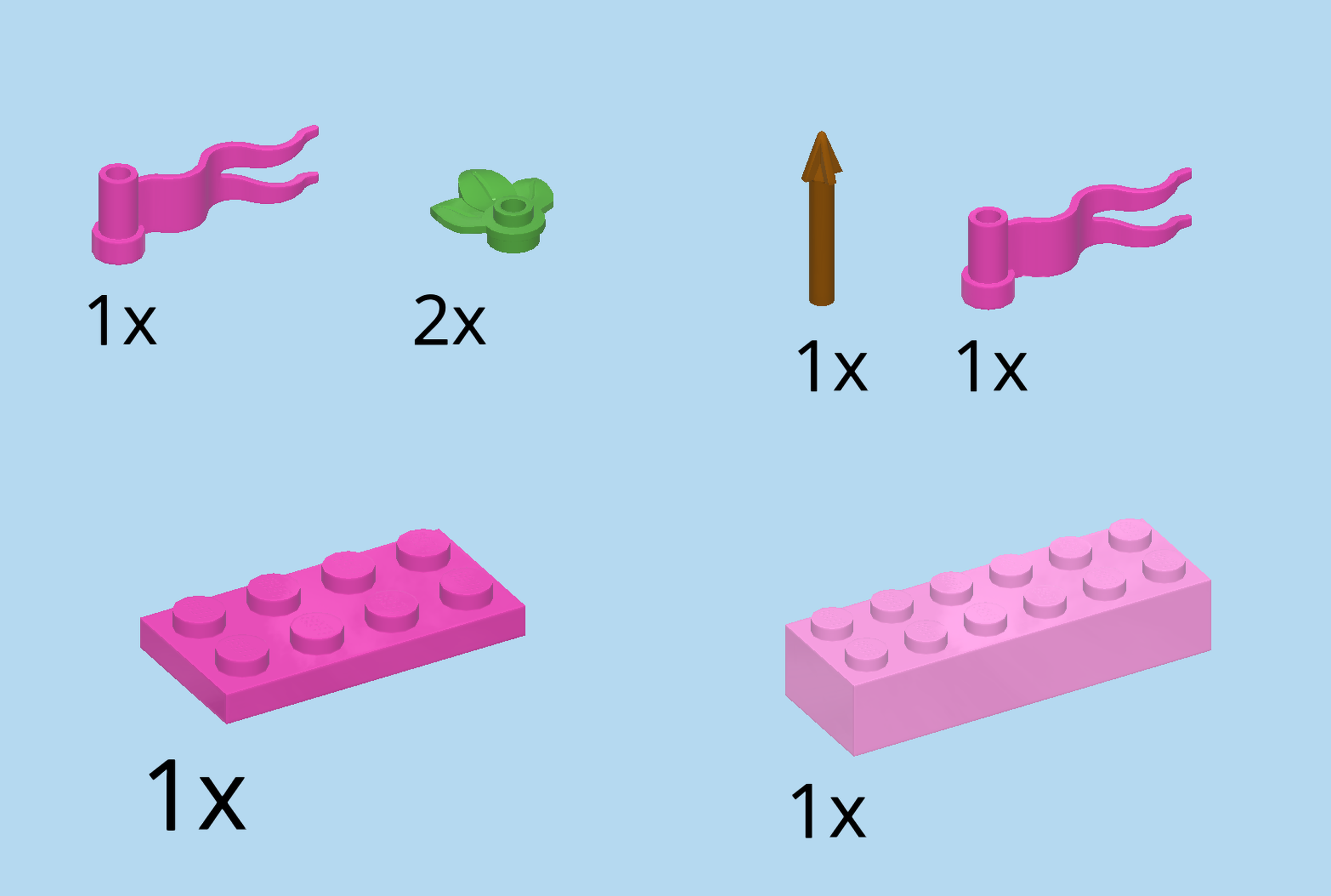}} \\
        \cmidrule(lr){1-2}
        \textbf{Reference} & 
        1 light purple brick [58,15,92,35]& \\
        \toprule[1.02pt]
        \textbf{mPLUG-OWL2} & 
        \multicolumn{2}{p{4cm}}{1 light purple brick[125,48,45,97]} \\
        \cmidrule(lr){2-3}
        Fine-tuned & 
        \multicolumn{2}{p{4cm}}{1 light purple brick [60,24,69,35]} \\ 
        \cmidrule(lr){1-3}
        \textbf{BLIP2} & 
        \multicolumn{2}{p{4cm}}{1 light purple brick [57,15,82,35] and 1 dark purple brick 2x6 [82,25,93,36]} \\
        \cmidrule(r){2-3}
        Fine-tuned & 
        \multicolumn{2}{p{4cm}}{1 light purple brick [58,15,92,35]} \\
            \cmidrule(lr){1-3}
        \textbf{LLaVA} & 
        \multicolumn{2}{p{4cm}}{The light purple brick is located at the center of the image, surrounded by yellow and black bricks.} \\
        \cmidrule(lr){2-3}
        Fine-tuned & 
        \multicolumn{2}{p{4cm}}{1 light purple brick [62,15,87,35]} \\
        \cmidrule(lr){1-3}
        \textbf{QWEN-VL} & 
        \multicolumn{2}{p{4cm}}{[20,61,30,78]} \\
        \cmidrule(lr){2-3}
        Fine-tuned & 
        \multicolumn{2}{p{4cm}}{1 light purple brick[57,15,82,35]} \\
        \cmidrule(lr){1-3}
        \textbf{InstructiBLIP} & 
        \multicolumn{2}{p{6cm}}{1 light purple brick 2x6 [57,15,82,35]} \\
        \cmidrule(r){2-3}
        Fine-tuned & 
        \multicolumn{2}{p{4cm}}{1 light purple brick [58,15,92,35]}  \\
        \cmidrule(r){1-3}
        \textbf{MiniGPT-v2} & 
        \multicolumn{2}{p{4cm}}{1 light purple brick [62,20,88,40]} \\
        \cmidrule(r){2-3}
        Fine-tuned & 
        \multicolumn{2}{p{4cm}}{1 light purple brick [57,17,85,37]} \\
        \toprule[1.02pt]
        \textbf{Otter} & 
        \multicolumn{2}{p{4cm}}{The 1 light purple brick 2x6 is on the left side of the image.} \\ 
        \cmidrule(lr){1-3}
        \textbf{MiniGPT-4} & 
        \multicolumn{2}{p{4cm}}{Sure! Here are the positions of the light purple brick in the image: Xleft = 0, Ytop = 25, Xright = 30, Ybottom = 50.} \\
        \cmidrule(lr){1-3}
        \textbf{GPT-4o} & 
        from PIL import Image \# Load the image  image\_path =$\dots$
        \\
        \bottomrule[1.1pt]
    \end{tabular}
    \caption{An example of object detection results. The object is highlighted in light purple, and its position is highlighted in blue.}
    \label{tab:obj_ep}
\end{table}

\section{Examples in Data Quality Assessment}

\begin{figure}[htb!]
    \centering
    \includegraphics[width=\columnwidth]{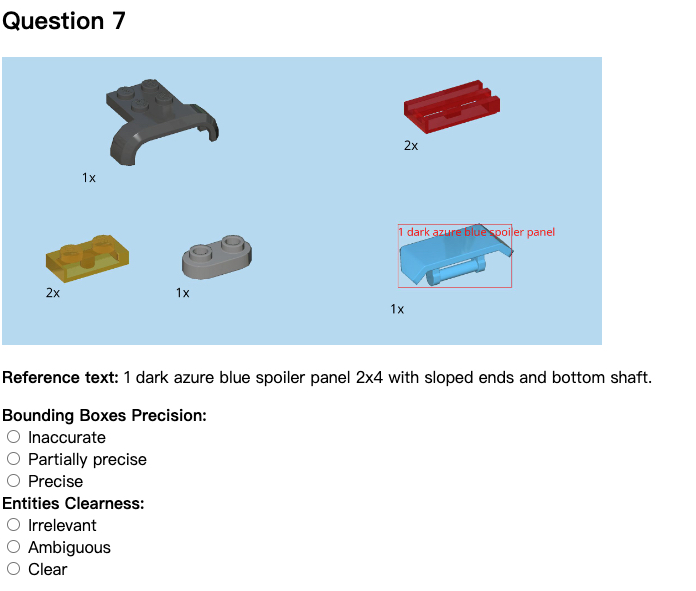}
    \caption{An example of evaluating data quality of task T2.}
    \label{fig:data_quality_t2}
\end{figure}

\begin{figure}[htb!]
    \centering
    \includegraphics[width=0.95\columnwidth]{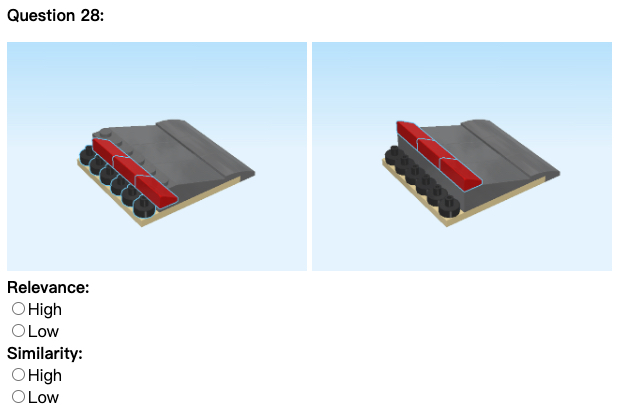}
    \caption{An example of evaluating data quality of task T2.}
    \label{fig:data_quality_t3}
\end{figure}

Figure \ref{fig:data_quality_t2} and \ref{fig:data_quality_t3} illustrate examples of the quality assessment for generated data in object detection (T2) and state detection (T3), respectively. 
We randomly selected 100 data examples per task to ensure a fair assessment.
Three annotators were provided with comprehensive guidelines on how to evaluate each dataset based on multiple criteria, accompanied by relevant examples. 
The first five data points mentioned in the guidelines were used as a sanity check to ensure that the annotators correctly understood and applied the evaluation criteria.
If annotators fail the sanity check, they undergo retraining until they pass. 
We evaluate annotations using the average scores of four criteria (i.e., entity disambiguity, boundary precision, state relevance, and state identifiability) and calculate Kappa values to assess annotation agreement.

\clearpage
\end{document}